\documentclass[letterpaper,10pt,journal,twoside]{IEEEtran}  

\IEEEoverridecommandlockouts                              

\ifdefined\pdfminorversion
\fi

\usepackage{amssymb}
\usepackage{amsmath} 
\usepackage{graphicx}
\usepackage{booktabs}
\usepackage{array}
\usepackage{cite}
\usepackage[caption=false,font=normalsize,labelfont=sf,textfont=sf]{subfig}
\usepackage{multirow} 
\usepackage{tabularx}
\usepackage{ragged2e}
\usepackage{threeparttable}
\usepackage{stfloats}
\usepackage{mathtools} 
\usepackage{xcolor}
\usepackage{tcolorbox}
\usepackage{url}

\title{Decentralized End-to-End Multi-AAV Pursuit Using Predictive Spatio-Temporal Observation via Deep Reinforcement Learning}

\author{Yude Li, Zhexuan Zhou, Huizhe Li, Yanke Sun, Yenan Wu, Yichen Lai,\\ Yiming Wang, Youmin Gong$^{*}$, and Jie Mei$^{*}$%
\thanks{Manuscript received: December 30, 2025; Revised: March 31, 2026; Accepted: May 18, 2026.}
\thanks{This paper was recommended for publication by Editor Giuseppe Loianno upon evaluation of the Associate Editor and Reviewers' comments.}
\thanks{This work was supported by Shenzhen Science and Technology Program (JCYJ20240813104923032), SAST Industry-University-Research Cooperation Fund (SAST2024-016), and  National Key Laboratory of Space Intelligent Control (HTKJ2024KL502003). (Corresponding authors: Youmin Gong and Jie Mei).}
\thanks{The authors are with  are with the School of Intelligence Science and Engineering, the Guangdong Key Laboratory of Intelligent Morphing Mechanisms and Adaptive Robotics, and the Shenzhen Key Lab for Advanced Motion Control and Modern Automation Equipments, Harbin Institute of Technology, Shenzhen, Guangdong 518055, China (e-mail: jmei@hit.edu.cn; gongyoumin@hit.edu.cn)}
\thanks{Digital Object Identifier (DOI): see top of this page.}
}

\begin{document}

\maketitle


\begin{abstract}
Decentralized cooperative pursuit in cluttered environments is challenging for autonomous aerial swarms, especially under partial and noisy perception. Existing methods often rely on abstracted geometric features or privileged ground-truth states, and therefore sidestep perceptual uncertainty in real-world settings. We propose a decentralized end-to-end multi-agent reinforcement learning (MARL) framework that maps raw LiDAR observations directly to continuous control commands. Central to the framework is the Predictive Spatio-Temporal Observation (PSTO), an egocentric grid representation that aligns obstacle geometry with predictive adversarial intent and teammate motion in a unified, fixed-resolution projection. Built on PSTO, a single decentralized policy enables agents to navigate static obstacles, intercept dynamic targets, and maintain cooperative encirclement. Simulations demonstrate that the proposed method achieves superior capture efficiency and competitive success rates compared to state-of-the-art learning-based approaches relying on privileged obstacle information. Furthermore, the unified policy scales seamlessly across different team sizes without retraining. Finally, fully autonomous outdoor experiments validate the framework on a quadrotor swarm relying on only onboard sensing and computing. Project details are available at \url{https://hitsz-mas.github.io/psto-aav-pursuit/}. 
\end{abstract}
\vspace{-1mm}

\vspace{-2mm}

\begin{figure}[!t] 
\centering  
\includegraphics[width=0.35\textwidth]{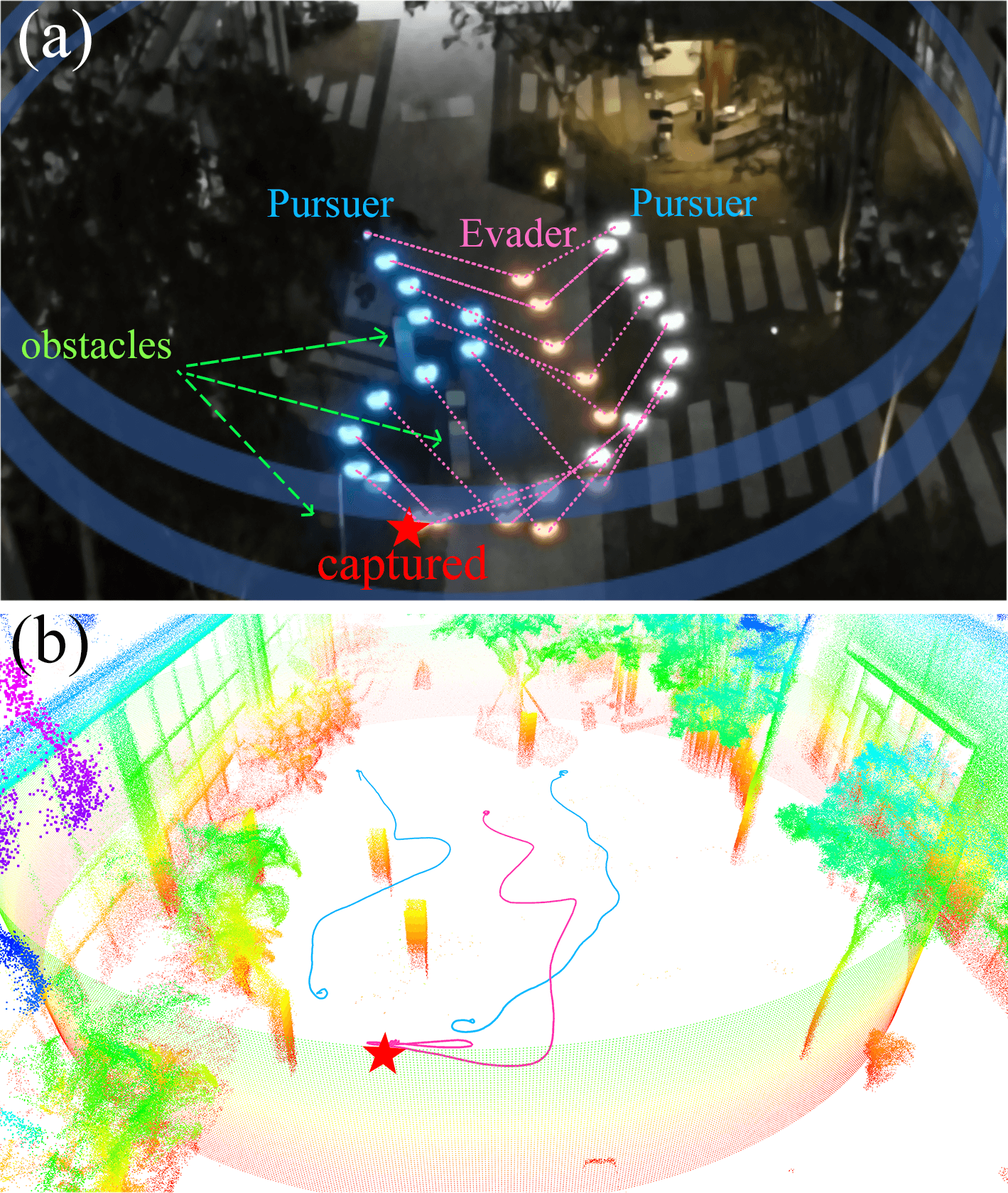}  
\caption{{Validation in an unstructured outdoor environment.} 
(a) Time-lapse of a fully autonomous 2-vs-1 pursuit. Pursuers (blue) execute a {coordinated pincer maneuver} to encircle the Evader (orange) among physical obstacles within a 9.0~m virtual boundary (blue ring). Dashed lines visualize the formation geometry at synchronized timestamps.
(b) Global state visualization. Trajectories and perception data visualized in the global frame. Note the {artificial point cloud wall} superimposed to enforce arena constraints.}
\label{main} 
\end{figure}
\vspace{-2mm}

\section{INTRODUCTION}

\IEEEPARstart{D}{ecentralized} aerial swarms promise capabilities in rapid exploration~\cite{zhou2023racer}, agile formation flight~\cite{zhou2022swarm}, and proximal cooperation~\cite{cao2025proximal}. Realizing such autonomy in unstructured environments is challenging: each agent must act from partial, egocentric observations under sensing noise and occlusion. Extracting relative state and cooperative intent directly from raw sensor data is therefore a central difficulty that traditional decentralized planners~\cite{hou2025primitive} typically avoid by assuming accurate, low-dimensional state estimates.

These challenges become more severe in dynamic, adversarial tasks. In the pursuit–evasion (P–E) setting, a team of pursuers must cooperatively intercept evaders~\cite{chung2011search}. Classical approaches based on differential games~\cite{isaacs1999differential} or simplified 2D grid worlds~\cite{vidal2002probabilistic} usually assume ideal sensing and abstract dynamics, and thus do not fully capture the partial observability and nontrivial vehicle constraints of real aerial swarms. 

Modern Multi-Agent Reinforcement Learning (MARL) has enabled sophisticated coordination, yet most methods still adopt a state-to-control paradigm. Policies are driven by processed, low-dimensional state vectors and hand-crafted geometric features, often relying on privileged knowledge of obstacle geometry or global evader states~\cite{zhang2022multi, chen2025online, liu2024game, qu2023pursuit}. Such formulations implicitly assume that an upstream perception module has already produced clean, fully observable states, thereby sidestepping the core perception-to-control problem: jointly interpreting high-dimensional, noisy sensor streams and coordinating actions in real time.

By contrast, end-to-end perception-to-control policies have achieved strong performance for single Autonomous Aerial Vehicle (AAV). Deep reinforcement learning has proven capable of mapping raw sensor inputs directly to agile maneuvers, as demonstrated across various complex tasks ranging from autonomous racing to navigation in unstructured environments~\cite{kaufmann2023champion, xu2025navrl, lu2024you, zhang2025learning, xu2024omnidrones}. Building on this, we extend the paradigm to decentralized multi-agent systems, addressing the specific challenges of joint perception and credit assignment.

In this work, we develop a decentralized end-to-end MARL framework that maps raw LiDAR observations directly to continuous control commands for multi-AAV pursuit. The key idea is the Predictive Spatio-Temporal Observation (PSTO), a unified egocentric predictive grid that jointly encodes dense obstacle geometry and the anticipated motion trajectories of the evader and teammates via a fixed-resolution spherical projection. Built on PSTO, a dual-stream convolutional policy learns to simultaneously reason about collision avoidance, interception, and multi-agent coordination without sparse geometric abstractions or global information, {as demonstrated in Fig. \ref{main}}.  {However, to facilitate cooperative maneuvers, pursuers within this decentralized setup explicitly share their local kinematic states via a communication network. This hardware-aligned assumption allows us to rigorously verify the cooperative strategy in real-world environments while isolating it from onboard relative perception errors.}

The key contributions are summarized as follows:
\begin{itemize}
    \item \textbf{Predictive Spatio-Temporal Observation (PSTO):} This unified representation bridges sparse predictive intent and dense environmental geometry via spatial alignment. By projecting raw LiDAR streams, adversarial forecasts, and teammate motions onto a fixed-resolution grid, it enables end-to-end learning, allowing the policy to implicitly reason about collision-free cooperative maneuvers  in cluttered environments.

    \item \textbf{Scalability across Team Sizes:} Leveraging the input-invariant nature of PSTO, the proposed unified policy can coordinate teams of varying sizes. It scales seamlessly across different team compositions without retraining, maintaining robust coordination performance.

    \item \textbf{Infrastructure-Free Swarm Deployment:} Fully autonomous outdoor flight experiments validate the complete framework. A quadrotor swarm equipped solely with on-board LiDAR sensing and on-board computing achieves zero-shot 2-vs-1 pursuit in unstructured environments.

\end{itemize}

\begin{figure*}[!t] 
\centering  
\includegraphics[width=0.82\textwidth]{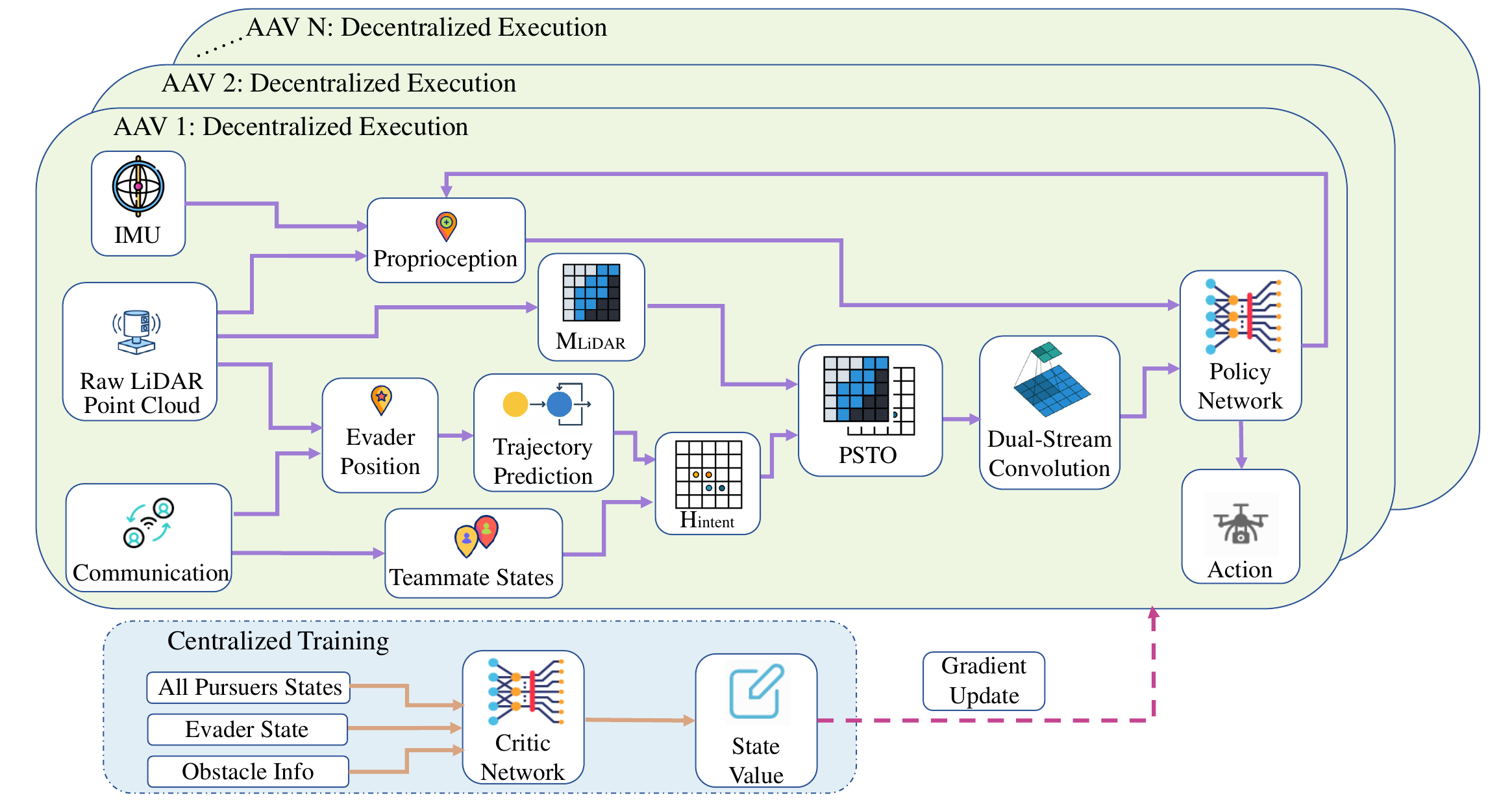}  
\caption{System overview of the proposed decentralized end-to-end pursuit framework. The architecture follows the CTDE paradigm. (Top) During decentralized execution, each pursuer generates the Predictive Spatio-Temporal Observation (PSTO), which is fused with proprioceptive data by the policy network to output continuous control commands. (Bottom) During centralized training, a critic network utilizes global state information to guide the policy optimization.}
\label{Fig.main2} 
\end{figure*}

\section{RELATED WORK}
The pursuit-evasion (P-E) problem~\cite{chung2011search} has been addressed by a wide spectrum of non-learning methods. These range from theoretical frameworks rooted in differential game theory~\cite{isaacs1999differential} and hierarchical grid decompositions~\cite{vidal2002probabilistic}, to geometric solutions like Voronoi partitions~\cite{tian2021distributed} or time-to-intercept calculations~\cite{pierson2016intercepting}, and finally to reactive heuristics such as Artificial Potential Fields (APF)~\cite{khatib1986real} and bio-inspired collectives~\cite{angelani2012collective, janosov2017group}.

However, these approaches share a critical vulnerability: they predominantly rely on the assumption of {precise, noise-free state knowledge}~\cite{tian2021distributed, pierson2016intercepting}. Such a reliance severely compromises the robustness of these approaches in real-world scenarios in the presence of sensory uncertainty and occlusion. Furthermore, rigorous solvers based on the {Hamilton-Jacobi-Isaacs (HJI) equation}~\cite{isaacs1999differential} suffer from the {curse of dimensionality}, while heuristics like APF are prone to {local minima}~\cite{khatib1986real,koren1991potential} and often neglect complex vehicle dynamics.

The rise of MARL has enabled complex decentralized policies, but many frameworks still rely on simplifications that prevent end-to-end autonomy. Hybrid methods that tune APF parameters with MARL~\cite{zhang2022multi} often assume {global, persistent access to the evader's position}. Even state-of-the-art MARL systems with strong coordination~\cite{de2021decentralized}, communication~\cite{wang2020cooperative}, and prediction~\cite{chen2025online, liu2024game} widely adopt a {state-to-control} paradigm. Policies are driven by processed, low-dimensional state vectors and hand-crafted representations, such as pre-computed relative angles and distances~\cite{qu2023pursuit}. While techniques like parameter sharing~\cite{gupta2017cooperative} have improved multi-agent scalability, these methods fundamentally rely on the abstracted geometric representations that are difficult to estimate reliably in real-time, or explicitly depend on ground-truth simulator states during training.

While significant progress has been made in end-to-end single-agent navigation~\cite{xu2025navrl, lu2024you, zhang2025learning, xu2024omnidrones}, these works do not address the inter-agent coordination and adversarial dynamics inherent in swarm pursuit. To our knowledge, the proposed method is the first multi-AAV framework to learn a decentralized P-E policy directly from LiDAR data, effectively bridging the gap between single-agent perception and multi-agent coordination.

\section{METHODOLOGY}

The overall architecture of the proposed decentralized end-to-end pursuit framework is illustrated in Fig. \ref{Fig.main2}. The system is grounded in the Decentralized Partially Observable Markov Decision Process (Dec-POMDP) formulation established in Sec. \ref{sec:problem_formulation}. As depicted in the decentralized execution pipeline shown at the top, each agent integrates raw LiDAR point clouds, IMU data, and shared teammate states to construct the unified Predictive Spatio-Temporal Observation (PSTO). The generation of this core representation, {which} spatially aligns dense geometry with sparse intent, is detailed in Sec. \ref{sec:PSTO_representation}. Subsequently, the observation is processed by a dual-stream backbone and concatenated with proprioceptive states to generate continuous control actions as described alongside the curriculum learning strategy in Sec. \ref{sec:policy_architecture}. Finally, the centralized training module shown at the bottom employs a critic network with global state access to guide the policy optimization.

\subsection{Problem Formulation as Dec-POMDP}
\label{sec:problem_formulation}

The setting is a planar multi-pursuer versus single-evader task in a cluttered arena, where each pursuer generates body-frame velocity commands based on local partial observations.

The  Decentralized Partially Observable Markov Decision Process (Dec-POMDP) is defined by the tuple $\langle \mathcal{N}, \mathcal{S}, \mathcal{A}, P, R, \Omega, O, \gamma \rangle$ ~~\cite{bernstein2002complexity}. Here, $\mathcal{N} = \{1, \dots, N\}$ is the set of pursuer agents, $\mathcal{S}$ is the global state space, and $\mathcal{A} = \prod_{i \in \mathcal{N}} \mathcal{A}_i$ denotes the joint action space, where $\mathcal{A}_i \subset \mathbb{R}^2$ represents parameters for Pursuer $i$'s high-level velocity command. $R$ is the shared reward function and $\Omega = \prod_{i \in \mathcal{N}} \Omega_i$ is the joint observation space. The local observation $o_i^t \in \Omega_i$ for Pursuer $i$ consists of its proprioceptive state $s_{\text{proprio}, i}^t \in \mathbb{R}^{12}$ and its Predictive Spatio-Temporal Observation (PSTO) $o_{\text{PSTO}}^i \in \mathbb{R}^{2 \times V_d \times H_d}$, {where $V_d$ and $H_d$ denote the fixed vertical and horizontal grid resolutions, respectively} (detailed in Sec.~\ref{sec:PSTO_representation}). Specifically, the proprioceptive state vector $s_{\text{proprio}, i}^t$ comprises the body attitude quaternion $\mathbf{q} \in \mathbb{R}^4$, body-frame linear velocity $\mathbf{v} \in \mathbb{R}^3$, angular velocity $\boldsymbol{\omega} \in \mathbb{R}^3$, and the previous control action $\mathbf{a}^{t-1} \in \mathbb{R}^2$. $O$ maps global state $s_t$ to the joint observation $o_t = (o_1^t, \dots, o_N^t)$, and $\gamma$ is the discount factor.

The objective is to learn decentralized policies $\pi_i(a_i^t | o_i^t)$ for each Pursuer $i$, sharing parameters $\theta$, that maximize the expected cumulative discounted reward:

\begin{equation} 
\label{eq:objective_final} 
J(\theta) = \mathbb{E}\left[\sum_{k=0}^{\infty} \gamma^k r_{t+k}\right],
\end{equation}
where $r_t = R(s_t, a^t)$ is the shared reward at time $t$.

\subsection{Predictive Spatio-Temporal Observation Representation}
\label{sec:PSTO_representation}

\begin{figure*}[!t] 
\centering  
\includegraphics[width=0.73\textwidth]{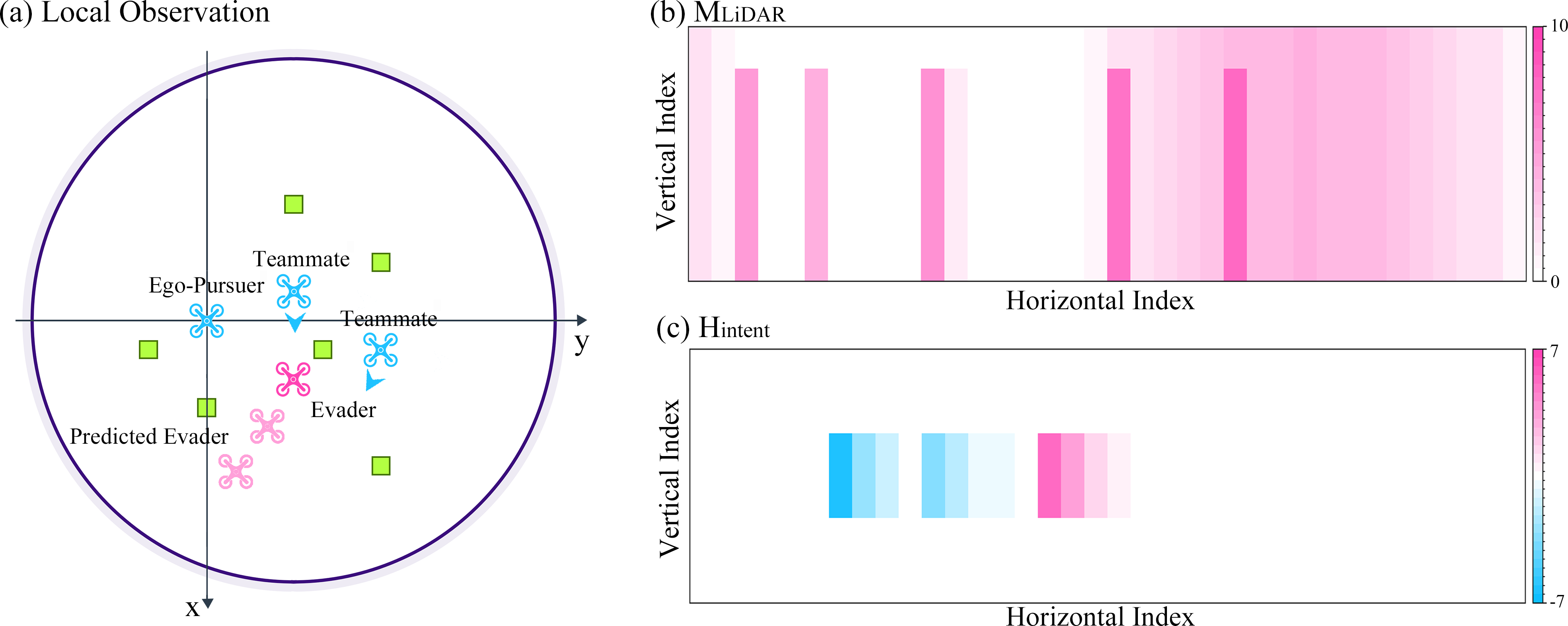}  
\caption{Visualization of the PSTO generation. (a) Egocentric View: The local environment observed by the Ego-Pursuer (center) in its body frame. It perceives the predicted Evader (pink), Teammates (blue), and static Obstacles (green) relative to itself. (b) \& (c) PSTO Channels: The corresponding PSTO generated by the Ego-Pursuer. (b) shows the obstacle depth map projected from the green squares in (a), and (c) shows the intent heatmap combining attraction to the Evader and repulsion from Teammates.}
\label{Figheat} 
\end{figure*}

To bridge the gap between geometric perception and cooperative intent, the {Predictive Spatio-Temporal Observation (PSTO)} is constructed as visualized in Fig. \ref{Figheat}. Unlike standard disjoint representations, PSTO {spatially aligns} sparse predictive signals with dense environmental constraints into a unified egocentric spherical grid. Formally, PSTO is defined as a two-channel tensor $o_{\text{PSTO}} \in \mathbb{R}^{2 \times V_d \times H_d}$: one channel encodes dense obstacle proximity $M_{\text{LiDAR}}$, while the other is the {predictive intention heatmap} $H_{\text{intent}}$, which aggregates the evader map $H_{\text{Evader}}$ and teammate map $H_{\text{Teammate}}$ via this shared projection.

\subsubsection{Mapping 3D Body-Frame Points to 2D Grid Coordinates}
\label{par:mapping_to_grid}
The transformation maps any 3D point ${}_{\mathcal{B}_i}p \in \mathbb{R}^3$ from Pursuer $i$'s body frame $\mathcal{B}_i$ to a pixel coordinate $(v, h)$ on the $V_d \times H_d$ pseudo-image grid.
This process, denoted by the function $\Pi: \mathbb{R}^3 \rightarrow \{0..V_d-1\} \times \{0..H_d-1\}$, follows the spherical projection method detailed in~\cite{li2025agile}.
It involves converting ${}_{\mathcal{B}_i}p$ to spherical coordinates $(r, \phi, \theta)$, {where $\phi$ and $\theta$ denote the azimuth and elevation angles respectively}, and quantizing them into grid indices $(v, h)$.
This quantization is based on the grid's field of view, defined by its minimum angles $(\phi_{\min}, \theta_{\min})$ and angular resolutions $(\Delta\phi, \Delta\theta)$:
\begin{equation} \label{eq:quantize_hv} 
h = \lfloor (\phi - \phi_{\min}) / \Delta\phi \rfloor, 
v = \lfloor (\theta - \theta_{\min}) / \Delta\theta \rfloor.
\end{equation}

In our implementation, the grid resolution is set to $V_d \times H_d = 120 \times 6$, covering a horizontal FOV of $360^\circ$ and a vertical FOV of $30^\circ$ ($[-10^\circ, 20^\circ]$). Points outside these bounds are disregarded.  {This vertical resolution maps agents at different altitudes to distinct rows ($v$), inherently preserving their 3D spatial separation in the grid.}

\subsubsection{Channel Generation} 

\paragraph*{(i) Physical Obstacle Proximity Map ($M_{\text{LiDAR}}$)}
The $M_{\text{LiDAR}}$ channel constructs a dense egocentric depth representation from the {current-frame} raw LiDAR point cloud $\mathcal{P}^t$. 
Let $\mathcal{S}_{v,h} = \{ p \in \mathcal{P}^t \mid \Pi(p)=(v,h) \}$ denote the set of points projecting onto the grid cell $(v, h)$.
To prioritize collision safety, a conservative projection strategy is employed:
\begin{equation} \label{eq:lidar_channel_assign_correct}
M_{\text{LiDAR}}(v, h) = 
\begin{cases}
    r_{\text{max}} - \min_{p \in \mathcal{S}_{v,h}} \|p\|_2, & \text{if } \mathcal{S}_{v,h} \neq \emptyset, \\
    0, & \text{otherwise}
\end{cases},
\end{equation}
where $r_{\text{max}}$ is the maximum sensing range of the LiDAR sensor. By aggregating points via the minimum range, the map explicitly preserves the closest environmental geometry. These ranges are then inverted relative to $r_{\text{max}}$ so that higher pixel intensities directly represent a higher collision threat, thereby facilitating feature extraction by the CNN backbone.

\paragraph*{(ii) Evader Component ($H_{\text{Evader}}$)}
The $H_{\text{intent}}$ channel encodes predictive adversarial intent. To capture the Evader's behavior and maintain prediction continuity under partial observability, an LSTM-based trajectory predictor is employed. The predictor first constructs the input history $H_e^t$ from the Evader's effective position (${}_{\mathcal{W}}p_{e, \text{eff}}^t$). This mechanism handles occlusion by defaulting to the prior prediction when the Evader is not observable:
\begin{equation} \label{eq:effective_pos}
{}_{\mathcal{W}}p_{e, \text{eff}}^t =
\begin{cases}
{}_{\mathcal{W}}\tilde{p}_e^t, & \text{if observable} \\
{}_{\mathcal{W}}\hat{p}_e^{t | t-1}, & \text{otherwise}
\end{cases},
\end{equation}
where ${}_{\mathcal{W}}\tilde{p}_e^t$ denotes the observed evader position, whose availability is assumed if observable by any pursuer, and ${}_{\mathcal{W}}\hat{p}_e^{t | t-1}$ represents the position predicted at the previous step. The LSTM $\mathcal{L}$ then predicts future world-frame waypoints over a prediction horizon $T_{\text{future}}$: $\{ {}_{\mathcal{W}}\hat{p}_e^{t+k | t} \}_{k=1}^{T_{\text{future}}} = \mathcal{L}(H_e^t)$. These predicted points, along with the current effective position, form the set ${}_{\mathcal{W}}\mathcal{P}_e^t$.
For Pursuer $i$, this set is transformed into the body frame ${}_{\mathcal{B}_i}\mathcal{P}_e^t$ via rigid body transformation:
\begin{equation} \label{eq:world_to_body}
{}_{\mathcal{B}_i}p = ({}_{\mathcal{W}}R_{\mathcal{B}_i}^t)^T (p - {}_{\mathcal{W}}p_i^t),
\end{equation}

For each predicted point ${}_{\mathcal{B}_i}p_k \in {}_{\mathcal{B}_i}\mathcal{P}_e^t$ corresponding to step $k$, we assign a value $\psi({}_{\mathcal{B}_i}p_k)$ to represent navigational attraction, decaying over both distance and time:
\begin{equation} \label{eq:evader_value_correct} 
\psi({}_{\mathcal{B}_i}p_k) = \frac{V_{\text{evader}}}{\|{}_{\mathcal{B}_i}p_k\|_2} \cdot \lambda_e^k, \quad k \in \{0, 1, \dots, T_{\text{future}}\} ,
\end{equation}
where $\lambda_e \in (0, 1]$ is the temporal decay factor and $V_{\text{evader}} > 0$ represents the target strength.

Finally, projecting these values onto the grid generates the heatmap.
Let $\mathcal{E}_{v,h} = \{ p \in {}_{\mathcal{B}_i}\mathcal{P}_e^t \mid \Pi(p)=(v,h) \}$ denote the subset of predicted points projecting onto the grid cell $(v, h)$.
Maximum aggregation is employed to retain the strongest attractive signal within each angular region:
\begin{equation} \label{eq:heatmap_evader_assign_clearer}
H_{\text{Evader}}(v, h) = 
\begin{cases}
    \max_{p \in \mathcal{E}_{v,h}} \psi(p), & \text{if } \mathcal{E}_{v,h} \neq \emptyset, \\
    0, & \text{otherwise}.
\end{cases}.
\end{equation}

\paragraph*{(iii) Teammate Component ($H_{\text{Teammate}}$)} 
Information about teammates is also projected onto the heatmap to encourage spatial separation. Pursuer $i$ receives the current world-frame ($\mathcal{W}$) state, including position ${}_{\mathcal{W}}p_j^t$ and velocity ${}_{\mathcal{W}}v_j^t$, from each teammate Pursuer $j$ ($j \neq i$) via communication. Teammate $j$'s future trajectory is predicted by linear extrapolation based on the current velocity:
\begin{equation} \label{eq:teammate_prediction}
{}_{\mathcal{W}}\hat{p}_j^{t+k | t} = {}_{\mathcal{W}}p_j^t + {}_{\mathcal{W}}v_j^t \cdot k \cdot \Delta t, \quad k \in [1, T_{\text{future}}],
\end{equation}
where $\Delta t$ is the timestep. The set of world-frame teammate points for projection is ${}_{\mathcal{W}}\mathcal{P}_j^t = \{ {}_{\mathcal{W}}p_j^t \} \cup \{ {}_{\mathcal{W}}\hat{p}_j^{t+k | t} \}_{k=1}^{T_{\text{future}}}$.  {Given the precise real-time velocities shared among teammates, this linear extrapolation provides highly accurate short-horizon predictions while avoiding the computational overhead and overfitting risks of sequence models \cite{8972605}.}

Following the transformation of predicted teammate trajectories into the body frame to form the union set $\mathcal{P}_{\text{team}} = \bigcup_{j \neq i} {}_{\mathcal{B}_i}\mathcal{P}_j^t$, each point $p_k \in \mathcal{P}_{\text{team}}$ corresponding to prediction step $k$ is assigned a value:
\begin{equation} \label{eq:teammate_value}
\psi(p_k) = -\frac{V_{\text{teammate}}}{\|p_k\|_2} \cdot \lambda_t^k, \quad k \in \{0, 1, \dots, T_{\text{future}}\},
\end{equation}
  {where $\lambda_t \in (0, 1]$ is the temporal decay factor and $V_{\text{teammate}} > 0$ denotes the social strength. Crucially, this temporal decay naturally mitigates any long-horizon deviation inherent in the constant-velocity assumption.}

The coordination map is generated by defining $\mathcal{T}_{v,h} = \{ p \in \mathcal{P}_{\text{team}} \mid \Pi(p)=(v,h) \}$ as the set of teammate points projecting onto grid $(v, h)$. 
Minimum aggregation captures the inter-agent constraints, thereby encouraging spatial separation and preventing overcrowding:
\begin{equation} \label{eq:heatmap_teammate_assign}
H_{\text{Teammate}}(v, h) = 
\begin{cases}
    \min_{p \in \mathcal{T}_{v,h}} \psi(p), & \text{if } \mathcal{T}_{v,h} \neq \emptyset, \\
    0, & \text{otherwise}.
\end{cases}.
\end{equation}

Finally, the complete intention heatmap is formed by summing the evader and teammate components:
\begin{equation} \label{eq:hintent_final_sum_correct}
H_{\text{intent}}(v, h) = H_{\text{Evader}}(v, h) + H_{\text{Teammate}}(v, h).
\end{equation}

\subsection{Policy Architecture and Training}
\label{sec:policy_architecture}

\subsubsection{Dual-Stream Convolutional Backbone: }
The two-channel PSTO, $o_{\text{PSTO}} = \text{Stack}(M_{\text{LiDAR}}, H_{\text{intent}})$, is processed by a Dual-Stream Convolutional Backbone. The LiDAR stream uses standard convolution and pooling to extract features from the dense $M_{\text{LiDAR}}$, while the Heatmap stream uses dilated convolutions~\cite{YuKoltun2016} and a Squeeze-and-Excitation (SE) based channel attention mechanism~\cite{hu2018squeeze} tailored to the sparse $H_{\text{intent}}$. The stream outputs are concatenated to form the final feature vector $f_{\text{PSTO}}$. Empirically, this decoupled design improves sample efficiency and final performance over a single shared convolutional stack.

\subsubsection{Actor-Critic Architecture}
The policy is trained using {Multi-Agent Proximal Policy Optimization (MAPPO)}~\cite{yu2022surprising} under the Centralized Training Decentralized Execution (CTDE) paradigm~\cite{lowe2017multi}, with a decentralized actor and a centralized critic. The actor $\pi_\theta$ computes the policy for agent $i$ from its local observation by feeding the concatenation of $f_{\text{PSTO}}$ and $s_{\text{proprio}, i}^t$ through an MLP whose final layer outputs the parameters $\alpha$ and $\beta$ of a Beta distribution $\mathcal{B}(\alpha, \beta)$, from which the continuous action $a_i^t$ is sampled~\cite{chou2017improving}. The critic $V_\phi$ utilizes the global state $s_t$ to guide optimization during training. In the deployment phase, each pursuer executes the {parameter-shared decentralized policy} using only its local PSTO and proprioceptive state.

\subsubsection{Reward Function}
The shared reward at time $t$ is composed of dense shaping terms and sparse terminal signals: $r_t = r_{\text{dense}} + r_{\text{sparse}}$. For brevity, time subscripts are omitted. We define the relative position vector as $\mathbf{r}_{ij} \triangleq \mathbf{p}_j - \mathbf{p}_i$ and the unit direction vector as $\mathbf{u}_{ij} \triangleq \mathbf{r}_{ij} / \|\mathbf{r}_{ij}\|$. The dense reward $r_{\text{dense}} = \sum w_k r_k$ consists of:

\begin{itemize}
    \item \textbf{Pursuit ($r_{\text{purs}}$):} Incentivizes rapid interception. Let $v_i = (\mathbf{v}_i - \mathbf{v}_e) \cdot \mathbf{u}_{ie}$ be the projected closing velocity. The reward balances average swarm progress with individual optimality:
$
        r_{\text{purs}} = \frac{1}{|\mathcal{N}|} \sum_{i \in \mathcal{N}} v_i + \min_{i \in \mathcal{N}}(v_i).
$
    
    \item \textbf{Coordination ($r_{\text{coord}}$):} Maintains formation scale via a Gaussian potential centered at the desired spacing $d_{\text{des}}$ with bandwidth $\sigma_{\text{coord}}$. Let $\bar{d}$ denote the average inter-agent distance among all pairs:
$
        r_{\text{coord}} = \exp\left(- {(\bar{d} - d_{\text{des}})^2}/{2\sigma_{\text{coord}}^2}\right).
$
    
    \item \textbf{Formation ($r_{\text{form}}$):} Adapted from~\cite{de2021decentralized}, this term promotes encirclement by maximizing the angular separation. Let $\mathbf{u}_{c,i}$ be the direction from agent $i$ to its closest teammate:
$
        r_{\text{form}} = \frac{1}{|\mathcal{N}|} \sum_{i \in \mathcal{N}} \arccos(\mathbf{u}_{c,i} \cdot \mathbf{u}_{ie}).
$
    
    \item \textbf{Safety ($r_{\text{obs}}$):} Penalizes proximity to obstacles using a log-barrier on the normalized minimum LiDAR reading $\hat{d}_i = \min(1, d_{\min,i}/d_{\text{safety}})$, which ensures zero penalty when safe and negative values otherwise:
$
        r_{\text{obs}} = \frac{1}{|\mathcal{N}|} \sum_{i \in \mathcal{N}} \log(\hat{d}_i).
$

    \item \textbf{Time Penalty ($r_{\text{time}}$):} A constant penalty $r_{\text{time}} = -1$ is applied at each step to incentivize minimum-time capture.
\end{itemize}

The sparse terminal reward $r_{\text{sparse}}$ assigns fixed values upon episode conclusion: \textbf{Capture} ($R_{\text{cap}}$) if $\min_i \|\mathbf{r}_{ie}\| < d_{\text{cap}}$; \textbf{Collision} ($R_{\text{coll}}$) if the agent collides with obstacles or teammates; \textbf{Escape} ($R_{\text{esc}}$) if the evader exceeds thresholds; or \textbf{Timeout} ($R_{\text{out}}$) if the episode duration exceeds $T_{\max}$.

\subsubsection{Progressive Training Curriculum}
To facilitate the learning of scalable coordination strategies and mitigate the sparse reward problem, a {progressive training curriculum}~\cite{bengio2009curriculum} is employed.
The training process gradually escalates {task difficulty} by increasing the evader's speed and obstacle density, while simultaneously decreasing the capture radius from a large initial value.
Additionally, the {team scale} is progressively expanded from a fundamental 2-vs-1 scenario to the final 4-vs-1 configuration.
This staged approach ensures that the policy masters basic interception skills before tackling complex multi-agent coordination, directly supporting the scalability across team sizes discussed in Sec. \ref{sec:results_scalability}.

\begin{table*}[!t]
\centering
\caption{Comparison of Ours (PSTO) against baselines in the 2-vs-1 scenario. SR: Success Rate ($\uparrow$); CT: Capture Time in steps ($\downarrow$). Best results are highlighted in bold.}
\label{tab:ours_first_comparison_final_corrected}
\resizebox{\textwidth}{!}{
\begin{tabular}{c c cc cc cc cc cc cc cc cc cc}
\toprule
 \multicolumn{2}{c}{\textbf{Conditions}} & \multicolumn{2}{c}{\textbf{Ours}} & \multicolumn{6}{c}{\textbf{Traditional Heuristics}}& \multicolumn{2}{c}{\textbf{SOTA (w/ GT)}} & \multicolumn{8}{c}{\textbf{Learning-Based Ablations}} \\
\cmidrule(lr){1-2} \cmidrule(lr){3-4} \cmidrule(lr){5-10} \cmidrule(lr){11-12} \cmidrule(lr){13-20}
 \multirow{2}{*}{\textbf{Speed}}&\multirow{2}{*}{\textbf{Obs.}} &  \multicolumn{2}{c}{PSTO}  & \multicolumn{2}{c}{Angelani~\cite{angelani2012collective}} & \multicolumn{2}{c}{APF~\cite{khatib1986real}} & \multicolumn{2}{c}{Janosov~\cite{janosov2017group}} & \multicolumn{2}{c}{OPEN~\cite{chen2025online}}
 & \multicolumn{2}{c}{{1-vs-1 Abl.}} & \multicolumn{2}{c}{Purely Reactive B.} & \multicolumn{2}{c}{Velocity-Aware B.} & \multicolumn{2}{c}{Separated-Input Abl.} \\
 \cmidrule(lr){3-4} \cmidrule(lr){5-6} \cmidrule(lr){7-8} \cmidrule(lr){9-10} \cmidrule(lr){11-12} \cmidrule(lr){13-14} \cmidrule(lr){15-16} \cmidrule(lr){17-18}\cmidrule(lr){19-20}
 & & SR & CT  & SR & CT & SR & CT & SR & CT & SR & CT & SR & CT & SR & CT & SR & CT & SR & CT\\
\midrule
\multirow{4}{*}{0.8} & 0 & {100.00\%} & 125.00 & 100.00\% & \textbf{105.80} & 100.00\% & 115.60 & 100.00\% & 131.10 & 100.00\% & 136.30 & 100.00\% & 133.90 & 100.00\% & 136.00 & 100.00\% & 137.20 & 100.00\% & 142.00 \\
& 3 & \textbf{100.00\%} & 122.13 & 97.22\% & {119.05} & 95.83\% & \textbf{116.47} & 93.06\% & 127.13 & 98.81\% & {129.20}  & 95.11\% & 128.88 & 88.19\% & 133.30 & 99.31\% & 136.95 & 97.92\% & 146.30 \\
& 6 & \textbf{99.31\%} & {116.88} & 90.28\% & 128.47 & 92.36\% & \textbf{115.38} & 89.58\% & 123.48 & 98.41\% & {129.13} & 96.53\% & 122.57 & 87.35\% & 129.47 & 97.92\% & 126.55 & 97.22\% & 142.05 \\
& 9 & 95.14\% & {114.55} & 84.63\% & 129.02 & 89.58\% & \textbf{112.97} & 81.94\% & 123.07 & \textbf{98.47\%} & {124.70} & 90.97\% & 121.95 & 85.96\% & 136.88 & {96.53\%} & 128.35 & 90.28\% & 143.30 \\
\midrule
\multirow{4}{*}{1.0} & 0 & {100.00\%} & 125.70 & 16.67\% & \textbf{88.00} & 0.00\% & - & 100.00\% & 157.00 & 100.00\% & 138.00 & 0.00\% & - & 79.17\% & 148.30 & 100.00\% & 139.80 & 100.00\% & 142.30 \\
& 3 & \textbf{99.31\%} & \textbf{125.17} & 41.67\% & 133.35 & 31.94\% & 138.78 & 90.28\% & 151.72 & 97.62\% & {131.70} & 47.55\% & 146.78 & 72.22\% & 145.22 & 96.53\% & 139.35 & 97.92\% & 143.18 \\
& 6 & {97.92\%} & \textbf{117.65} & 59.72\% & 146.05 & 48.28\% & 152.20 & 81.25\% & 147.38 & \textbf{98.02\%} & {132.31}  & 72.86\% & 136.27 & 81.79\% & 142.07 & 95.08\% & 130.13 & 93.75\% & 144.37 \\
& 9 & {95.83\%} & \textbf{118.93} & 63.89\% & 146.75 & 59.03\% & 137.47 & 70.14\% & 148.03 & \textbf{98.47\%} & {129.33} & 81.94\% & {134.73} & 79.68\% & 140.10 & 94.44\% & 130.27 & 90.97\% & 148.37 \\
\midrule
\multirow{4}{*}{1.2} & 0 & {100.00\%} & \textbf{125.90} & 0.00\% & - & 0.00\% & - & 66.67\% & 165.30 & 100.00\% & 137.80 & 0.00\% & - & 66.67\% & 144.40 & 100.00\% & 138.90 & 100.00\% & 142.60 \\
& 3 & \textbf{98.61\%} & \textbf{125.70} & 37.50\% & 151.13 & 36.81\% & 144.38 & 54.86\% & 170.58 & 98.21\% & 134.53 & 50.30\% & 145.92 & 68.06\% & 144.73 & 97.92\% & 139.87 & 97.92\% & 144.10 \\
& 6 & \textbf{100.00\%} & \textbf{121.00} & 48.91\% & 160.72 & 43.75\% & 148.52 & 55.56\% & 173.05 & 98.81\% &  135.70 & 69.26\% & 138.30 & 75.54\% & 146.52 & 93.69\% & 132.35 & 94.44\% & 144.55 \\
& 9 & {95.83\%} & \textbf{120.03} & 55.56\% & 147.67 & 54.17\% & 137.98 & 54.17\% & 175.00 & \textbf{97.96\%} & 131.53 & 75.69\% & 136.37 & 77.63\% & 143.27 & 93.75\% & 134.15 & 90.28\% & 146.65 \\
\midrule
\multirow{4}{*}{1.4} & 0 & {100.00\%} & \textbf{125.30} & 0.00\% & - & 0.00\% & - & 0.00\% & - & 100.00\% &  141.40 & 0.00\% & - & 62.50\% & 142.50 & 100.00\% & 139.50 & 100.00\% & 143.10 \\
& 3 & \textbf{98.61\%} & \textbf{126.32} & 37.50\% & 157.83 & 38.19\% & 153.05 & 20.14\% & 206.37 &  97.62\% & 137.12 & 48.25\% & 149.40 & 64.58\% & 146.10 & 98.61\% & 141.38 & 97.92\% & 144.43 \\
& 6 & {97.22\%} & \textbf{121.57} & 45.98\% & 156.50 & 42.60\% & 157.90 & 25.69\% & 212.42 & \textbf{98.41\%} & 139.10 & 67.69\% & 141.27 & 71.83\% & 146.03 & 93.60\% & 134.98 & 93.75\% & 144.15 \\
& 9 & {94.44\%} & \textbf{122.83} & 54.86\% & 157.27 & 49.31\% & 143.90 & 32.64\% & 191.32 & \textbf{98.98\%} & 134.44 & 72.22\% & 135.50 & 75.00\% & 146.77 & 95.83\% & 137.03 & 89.58\% & 146.75 \\
\midrule
\multirow{4}{*}{1.6} & 0 & {100.00\%} & \textbf{126.90} & 0.00\% & - & 0.00\% & - & 0.00\% & - & 100.00\% & 143.90 & 0.00\% & - & 79.17\% & 175.40 & 100.00\% & 140.10 & 95.83\% & 145.00 \\
& 3 & \textbf{97.92\%} & \textbf{126.88} & 38.89\% & 166.98 & 39.58\% & 161.85 & 15.28\% & 211.68 & 97.62\% & 141.78 & 47.55\% & 153.82 & 65.28\% & 158.90 & 97.22\% & 142.40 & 96.53\% & 146.43 \\
& 6 & \textbf{97.22\%} & \textbf{124.42} & 47.86\% & 167.05 & 43.03\% & 160.92 & 18.48\% & 206.72 & 96.83\% & 141.08 & 65.07\% & 142.50 & 72.55\% & 154.67 & 92.21\% & 137.50 & 93.75\% & 144.37 \\
& 9 & {95.14\%} & \textbf{122.67} & 54.53\% & 166.17 & 49.37\% & 150.00 & 21.23\% & 208.60 & \textbf{97.96\%} & 138.39 & 72.04\% & 137.15 & 75.54\% & 148.60 & 94.44\% & 140.55 & 91.58\% & 146.55 \\
\bottomrule
\end{tabular}
} 
\end{table*}

\section{RESULTS AND DISCUSSION}
\label{sec:results}

\subsection{Experimental Setup}
\label{sec:exp_setup}

Training is performed in {NVIDIA Isaac Sim} using the {OmniDrones} platform~\cite{xu2024omnidrones} on an RTX 5880 Ada GPU with 1024 parallel environments. 
 {To facilitate robust sim-to-real transfer, Domain Randomization (DR)~\cite{tobin2017domain} is applied during training by injecting Gaussian noise into the pursuers' proprioceptive states and relative observations.}
The control policy operates at $10$~Hz, interacting with a $100$~Hz physics simulation.
Experiments are conducted in a circular arena ($r=9.0$~m) populated with {randomly placed} static obstacles ($0.4\times0.4$~m) to ensure environmental diversity.
Pursuers are subject to kinematic constraints with a maximum speed of {$v_{\max}=0.8$~m/s} and a limited perception horizon of {$r_{\max}=10.0$~m}.
The {Evader} uses a specialized multi-modal APF policy with tangential vector fields, producing aggressive wall-following maneuvers that challenge the pursuers.
At the start of each episode, the Evader is initialized at a random position, and the Pursuer team is spawned $3.5$~m behind it, with all agents facing the positive x-axis.
A successful capture is recorded when any pursuer closes within $1.4$~m. 
Performance is quantified by two primary metrics:
\begin{itemize}
    \item \textbf{Success Rate (SR):} The percentage of episodes where the pursuers successfully capture the evader within the time limit ($T_{\max}=30.0$~s).
    \item \textbf{Capture Time (CT):} The average number of control steps required to achieve capture, calculated only over successful episodes.
\end{itemize}
Quantitative results (Table \ref{tab:ours_first_comparison_final_corrected}) are averaged over {150 independent rollouts} per condition.

\begin{figure*}[!t]
\centering
\includegraphics[width=0.88\textwidth]{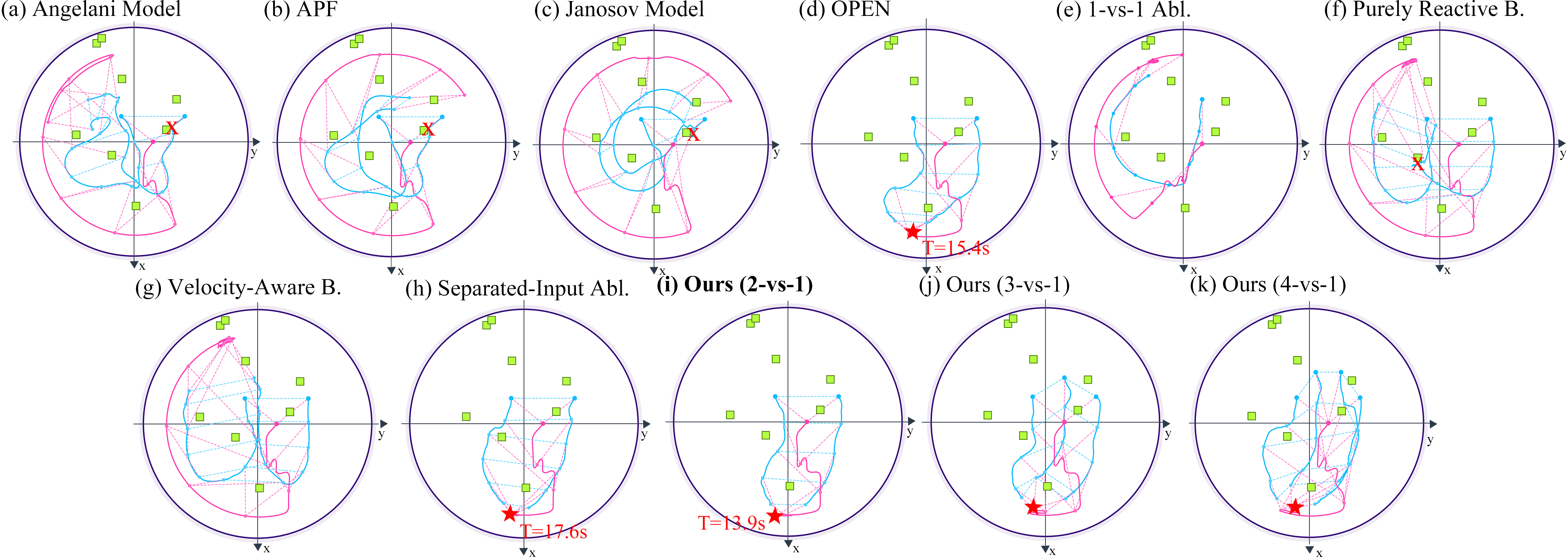}
\caption{Trajectory visualization ($1.6$~m/s, 8 obstacles).
\textbf{(a)--(c) Traditional Heuristics:} Collision (`$\times$') or escape due to poor coordination.
\textbf{(d) OPEN (SOTA)~\cite{chen2025online}:} Captures (`$\star$') but requires {privileged obstacle states}.
\textbf{(e)--(h) Ablations:} (e)--(g) Fail (speed deficit/containment loss) (h) Captures (`$\star$') but takes a {conservative detour}.
\textbf{(i)--(k) Ours (PSTO) \& Scalability:} (i) Executes rapid capture (`$\star$'). (j)--(k) The {unified policy} scales to 3-vs-1 and 4-vs-1, achieving {multi-angle containment}.
(Note: Dashed lines visualize synchronized formation geometry; baselines continue after collision to show trends.)}
\label{FigTrajectory}
\end{figure*}

\subsection{Comparison with Traditional Methods}
\label{sec:results_traditional}

We compare against three heuristics (APF~\cite{khatib1986real}, Angelani~\cite{angelani2012collective}, Janosov~\cite{janosov2017group}), all optimized at $0.8$~m/s. However, their performance degrades rapidly at higher speeds. Reactive baselines suffer from {oscillations}, while the predictive Janosov model remains purely reactive~\cite{janosov2017group}, treating obstacles merely as simple repulsive constraints.  As shown in {Fig. \ref{FigTrajectory}(a)--(c)}, these deficiencies lead to collisions or escapes. Notably, the rise in baseline success rates in dense clutter is an {environmental artifact} caused by obstacles trapping the evader. In contrast, {Ours (PSTO)} demonstrates superior robustness. By unifying intent and geometry, it maintains success rates above {94\%} across all conditions without retuning, achieving active interception. Crucially, the policy learns to exploit the arena boundary for cooperative entrapment, enabling the slower swarm to corner high-speed evaders.

\subsection{Efficacy of the PSTO Representation}
\label{sec:results_rl_ablation}

\subsubsection{Baselines and Experimental Setup}
\label{sec:def_baselines}
To rigorously evaluate the efficacy of dense geometric representations and isolate the contributions of our perception backbone, we compare {Ours (PSTO)} against a state-of-the-art (SOTA) method and four learning-based ablations. To ensure fairness, all variants were trained using identical MAPPO hyperparameters.

\begin{itemize}
    \item \textbf{OPEN (SOTA)~\cite{chen2025online}:} 
    Representing the current SOTA in multi-UAV pursuit, this baseline employs an evader prediction-enhanced network (LSTM) and a multi-head self-attention mechanism. {Unlike our vision-based approach, OPEN relies on privileged ground-truth obstacle states} rather than onboard sensors, and directly uses teammate states as vector inputs.
    \item \textbf{1-vs-1 Ablation:} A control experiment evaluating a single pursuer against the evader under identical environmental constraints. This setup serves to isolate individual tracking capabilities from cooperative behaviors.
    \item \textbf{Purely Reactive Baseline (Intent-Blind):} An MLP-based policy that processes sparse local sensing rather than the dense $M_{\text{LiDAR}}$, without access to the predicted future trajectories of the evader or teammates. This architecture serves as a representative benchmark for state-to-control MARL approaches~\cite{yu2022surprising}.
    \item \textbf{Velocity-Aware Baseline (Vector Intent + MLP):} Extends the reactive baseline with ground-truth relative velocities of the evader and teammates. To ensure a fair comparison of intent awareness without sequence modeling overhead, true velocities are used as an ideal proxy for the LSTM module. Crucially, it retains the MLP backbone, processing {sparse geometric features} and intent as concatenated vectors.
    \item \textbf{Separated-Input Ablation (Vector Intent + CNN):} Processes the {dense raw sensor grid} $M_{\text{LiDAR}}$ via a {CNN backbone}, while treating {intent information} as a {disjoint vector} that is concatenated with the extracted visual features, lacking spatial alignment.
\end{itemize}

\subsubsection{Performance Analysis}
We analyze the behavioral divergence by synthesizing the statistical metrics in {Table \ref{tab:ours_first_comparison_final_corrected}} with the trajectory visualizations in {Fig. \ref{FigTrajectory}(d)-(i)}, focusing on how different encoding schemes couple intent and geometry.

\textbf{1) The Necessity of Cooperation and Intent:}
The results first establish that multi-agent cooperation is prerequisite; the single pursuer in the {1-vs-1 Ablation} fails physically against the faster evader ({Fig. \ref{FigTrajectory}(d)}). Furthermore, intent awareness is critical. The {Purely Reactive Baseline}, relying solely on instantaneous positions, lacks the capacity to anticipate the future motion of either the evader or teammates. Consequently, it is unable to execute proactive interception, resulting in a suboptimal success rate, trailing by over 15\% compared to intent-aware baselines in dense clutter.

\textbf{2) The Limitation of Disjoint Representations:}
While the {Velocity-Aware} and {Separated-Input} baselines achieve competitive success rates, they suffer from inefficiency as navigational intent and geometry remain decoupled. Specifically, by treating obstacles as isolated range readings (MLP) or intent as a detached vector (Separated-Input), these representations lack the spatial alignment required for tactical encirclement. Consequently, agents resort to passive trailing or conservative detours. In contrast, PSTO unifies intent and depth into a single grid. This projection-level coupling enables the policy to identify efficient shortcuts through clutter (Fig. \ref{FigTrajectory}(h)), reducing capture time by 16.3\% compared to the disjoint baselines.

\textbf{3) Comparative Analysis with SOTA:}
 {Representing the recent state-of-the-art in MARL-based pursuit,} the OPEN baseline employs a {multi-head attention mechanism} to process {privileged ground-truth states}~\cite{chen2025online}. While this architecture ensures competitive success rates, it inherently treats environmental constraints and navigational intent as disjoint semantic tokens. Lacking explicit spatial fusion, this abstraction obscures the feasibility of narrow gaps, necessitating conservative safety margins (Fig. \ref{FigTrajectory}(d)). In contrast, PSTO's spatially aligned projection enables agents to directly perceive and exploit traversable corridors. Consequently, our approach achieves a decisive advantage in agility, consistently reducing capture time by 10--15 steps.

\subsection{Scalability Analysis}
\label{sec:results_scalability}

To achieve scalability across team sizes, a progressive curriculum training strategy was employed. The policy was initialized in the 2-vs-1 setting, then sequentially trained in 3-vs-1 and 4-vs-1 scenarios.  {Crucially, the single final policy saved after the 4-vs-1 stage was directly evaluated across all team configurations without any parameter fine-tuning.}

As summarized in Table \ref{tab:scalability}, this approach demonstrates robust generalization.  {Specifically, when re-tested in the initial 2-vs-1 scenario, it exhibited no catastrophic forgetting or performance degradation.} Even at a high evader speed of $v = 2.0$ m/s, success rates remain above $89\%$ across all configurations. Notably, the Capture Time decreases monotonically as team size increases. This confirms that the PSTO heatmap allows the network to learn a generalized cooperative logic that aggregates spatial constraints from varying numbers of agents rather than overfitting to a specific count.

This scalability is visually corroborated in Fig. \ref{FigTrajectory}(i)-(k). While the 2-vs-1 configuration (i) successfully intercepts the evader, the 3-vs-1 (j) and 4-vs-1 (k) configurations leverage additional agents to accelerate capture. As observed in the trajectory evolution, agents originating from identical initial sectors exhibit adaptive spatial distribution depending on team composition. In the 4-vs-1 scenario, pursuers instinctively widen their attack angles compared to the 3-vs-1 case to maximize spatial coverage. By forming a multi-angle encirclement that rapidly compresses the evader's feasible space, the swarm achieves higher interception efficiency.

\begin{table}[!t]
    \centering
    \caption{Scalability of the PSTO policy across team sizes at $v = 2.0$ m/s. SR: success rate; CT: capture time (steps).}
    \label{tab:scalability}
    \footnotesize
    \begin{tabular}{c c c c}
        	\toprule
        Obs. & Team Size & SR & CT \\ 
        \midrule
        \multirow{3}{*}{0} & 2-vs-1 & 100.00\% & 129.50 \\
                             & 3-vs-1 & 100.00\% & 110.20 \\
                             & 4-vs-1 & 100.00\% & 107.90 \\
        \midrule
        \multirow{3}{*}{3} & 2-vs-1 & 95.00\% & 127.22 \\
                             & 3-vs-1 & 93.30\% & 108.94 \\
                             & 4-vs-1 & 91.49\% & 101.17 \\
        \midrule
        \multirow{3}{*}{6} & 2-vs-1 & 91.67\% & 125.24 \\
                             & 3-vs-1 & 91.52\% & 108.92 \\
                             & 4-vs-1 & 89.84\% & 102.97 \\
        \bottomrule
    \end{tabular}
\end{table}

\begin{figure}[!t] 
\centering 
\includegraphics[width=0.36\textwidth]{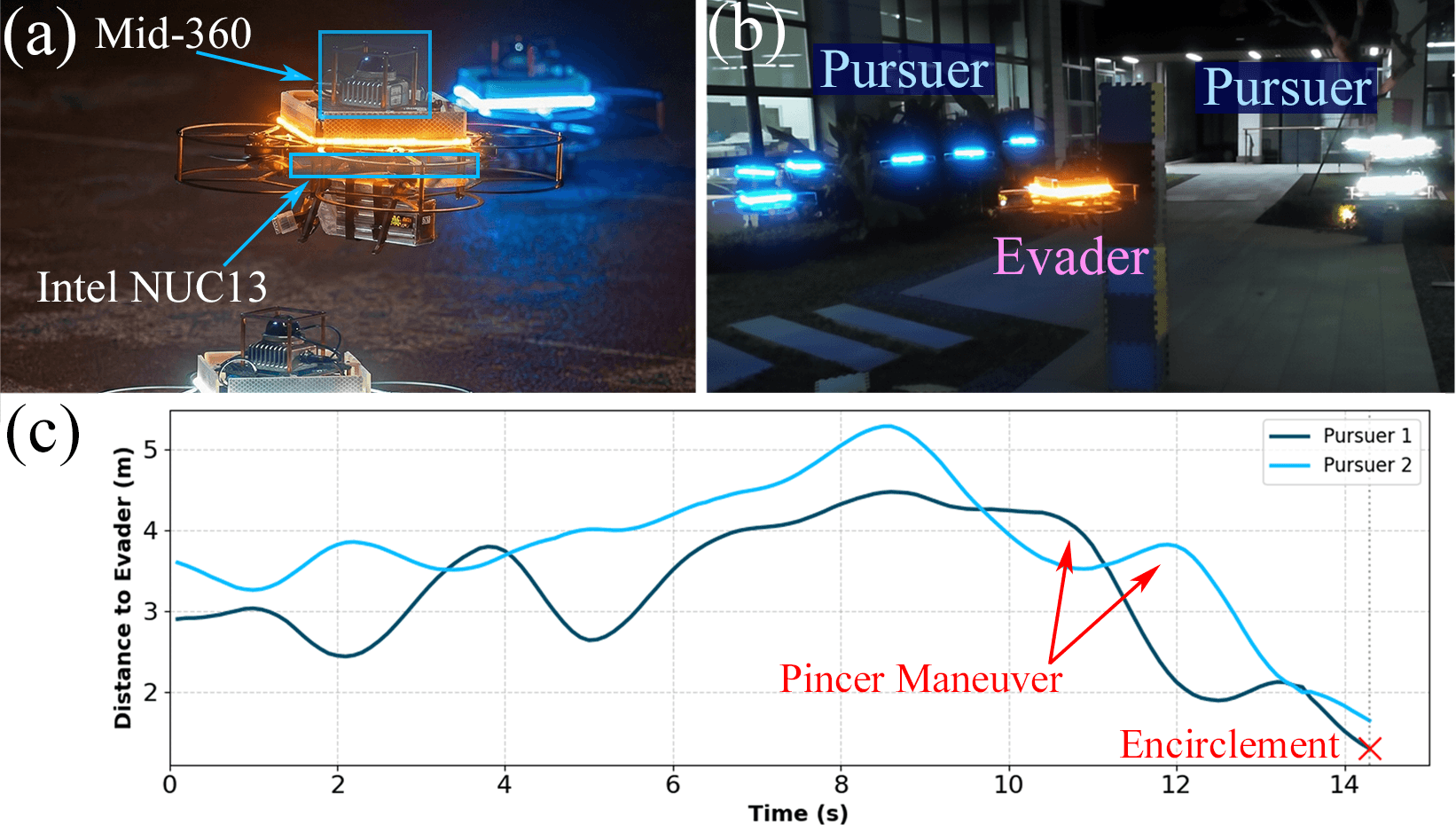} 
\caption{{Fully autonomous system and outdoor validation.} 
{(a)} The quadrotor equipped with Livox Mid-360 and NUC 13.
{(b)} Swarm pursuit in an unstructured outdoor environment. 
{(c)} Quantitative results {corresponding to the mission visualized in Fig. \ref{main}}. The distance evolution reveals a { pincer maneuver}, culminating in successful {encirclement}.}
\label{exp} 
\end{figure}


\subsection{Physical Experiment} 
\label{sec:results_physical}

To validate the transferability of the proposed method, we deployed the PSTO-based policy on custom-built quadrotors in an unstructured outdoor environment. The system operates in a fully decentralized manner using only on-board LiDAR and computing resources (Intel NUC 13, Core i7-1360P; hardware details in Fig. \ref{exp}(a)). The onboard camera is used solely for visualization.  Real-time state estimation is achieved via the Swarm-LIO algorithm~\cite{zhu2024swarm}. 
The perception-to-control pipeline runs efficiently on the onboard computer with an average latency of $\sim$3.87 ms (Encoding: 2.61 ms, Inference: 1.26 ms), well within the 100 ms control cycle.

To replicate the confined training arena in the open outdoor field, a virtual circular boundary (radius 9.0~m) is superimposed onto the raw LiDAR observations as artificial obstacle points (as visualized in Fig. \ref{main}). Furthermore, as an experimental simplification to focus on policy validation, when the evader is observable, its relative state is derived directly from the shared Swarm-LIO estimates, serving as a reliable proxy for the on-board detection system.  {Furthermore, the system mitigates potential communication degradation in real-world deployments via Swarm-LIO's filtering and PSTO's predictive delay compensation (Eq. \ref{eq:effective_pos}).}

The learned policy transfers zero-shot to outdoor flights, where the pursuers achieve collision-free coordinated encirclement among static obstacles, as shown in Fig. \ref{main} and supported by the telemetry in Fig. \ref{exp}(c). These results indicate that PSTO captures geometric features robust enough for on-board deployment under real-world sensor noise.

\section{CONCLUSIONS}
\label{sec:conclusion}

This paper has presented a decentralized end-to-end MARL framework for multi-AAV pursuit, centered on the Predictive Spatio-Temporal Observation (PSTO). By spatially aligning raw LiDAR geometry with predictive adversarial intent and teammate motion, PSTO has enabled the mapping of high-dimensional sensory inputs directly to continuous control, eliminating the reliance on sparse geometric abstractions or privileged global information. Simulation results have demonstrated that the proposed method outperforms both heuristic and strong learning-based baselines, achieving superior capture efficiency compared to state-of-the-art controllers that rely on privileged obstacle maps. Furthermore, the framework has exhibited robust scalability across varying team sizes under a single unified policy. Fully autonomous outdoor experiments have further validated the feasibility and robustness of the system using solely onboard sensing and computation.

While the PSTO representation is inherently extensible to 3D environments, our current validation has focused on planar pursuit scenarios to isolate core coordination dynamics from the complexities of vertical perception. Additionally, we have relied on shared state estimates for evader tracking to prioritize the verification of the coordination policy. Future work will integrate fully onboard perception for evader detection and extend the framework to unrestricted 3D pursuit–evasion in volumetric environments.





\bibliographystyle{IEEEtran}
\bibliography{references}{}

\end{document}